\documentclass[article, 10pt, conference]{ieeeconf}  

\IEEEoverridecommandlockouts

\usepackage{graphicx}
\usepackage{epsfig}
\usepackage{mathptmx}
\usepackage{times}
\usepackage{amsmath}
\usepackage{amssymb}
\usepackage{booktabs}

\title{\LARGE \bf
Robot Navigation and Obstacle Avoidance from \\Randomly Positioned External Cameras
}

\author{Raul Steinmetz$^{1}$, Jair Augusto Bottega$^{1}$, Ayanori Yorozu$^{1}$, and Akihisa Ohya$^{1}$%
\thanks{*This work was supported by the Ministry of Education, Culture, Sports, Science and Technology (MEXT), Japan.}%
\thanks{$^{1}$All authors are with the University of Tsukuba, Tsukuba, Ibaraki 305--8573, Japan. %
{\tt\small raulsteinmetz08@gmail.com, jairaugustobottega@gmail.com, yorozu@cs.tsukuba.ac.jp, ohya@cs.tsukuba.ac.jp}}%
\thanks{Code, data, and video demonstrations are available at {\tt\small https://github.com/raulsteinmetz/vine}.}%
}

\begin{document}

\maketitle
\thispagestyle{empty}
\pagestyle{empty}

\begin{figure*}[!t]
    \centering
    \includegraphics[width=0.9\textwidth]{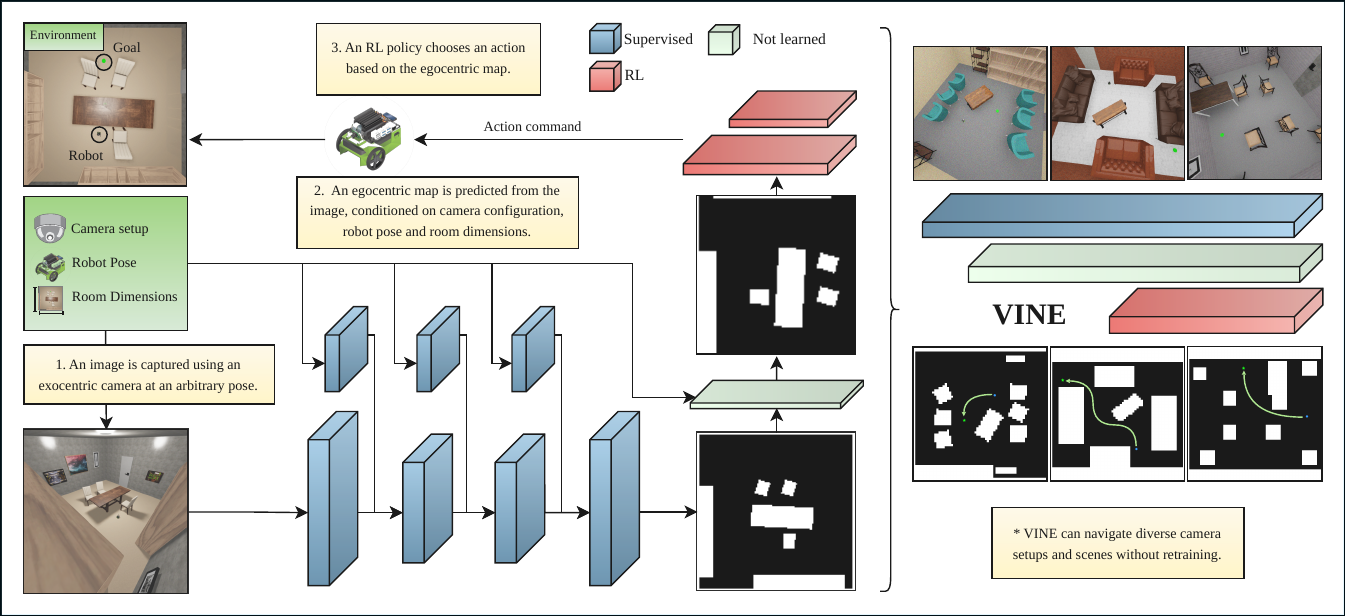}
    \caption{Overview of \textit{VINE}. A supervised-learned obstacle map estimation network (blue) predicts a global obstacle map from an exocentric image, conditioned on the approximate camera pose matrix and room width and depth. The figure shows one of the two options for the controller, in which the map is warped to the robot's egocentric frame (green) using the robot's pose and used by a reinforcement-learned policy (red) to select actions, while in the other option a planner such as A* acts directly on the global map. Within a camera setup, VINE handles randomized camera poses without retraining.}
    \label{fig:main}
\end{figure*}

\begin{abstract}

We study robot navigation in indoor environments from fixed exocentric cameras (infrastructure-mounted, external to the robot). We show that standard Proximal Policy Optimization (PPO), which succeeds with onboard cameras, fails to learn this navigation end-to-end in our setup. This failure results from the perspective mismatch between exocentric observations and egocentric action and goal formulations, which is too complex to resolve from the reward signal alone. To address this, we propose Viewpoint-Invariant Navigation from Exocentric Cameras (VINE). VINE uses an obstacle map estimation network, trained with supervised learning, that takes exocentric images, camera pose, robot pose, and room dimensions as input and produces a viewpoint-invariant obstacle map. This map can be used either by a PPO policy or by a planner such as A*. VINE's map estimation performs better than perspective mapping approaches, handles cameras placed across a wide range of positions and orientations without view-specific tuning, requires only approximate camera pose estimates (noise with a standard deviation of 30~cm per axis in $x$, $y$, and $z$ and about 20$^\circ$ of rotation error on average), is robust to intrinsic variation, and extends to two-camera setups with partial coverage. Despite operating from cameras randomized in position during evaluation, VINE surpasses the navigation success rate of PPO with an onboard camera in simulated environments.

\end{abstract}

\section{Introduction}

Navigation is a fundamental capability for autonomous mobile robots. Deep Reinforcement Learning (DRL) has emerged as a prominent approach for learning navigation policies directly from raw sensor data, yet nearly all prior work focuses on policies that act on onboard sensors, such as cameras and LiDAR mounted on the robot~\cite{zhu2021deep}. Learning to navigate from exocentric cameras, mounted in the environment rather than on the robot, remains largely unexplored, despite such sensors being ubiquitous in indoor environments. These cameras offer concrete advantages over onboard sensors for robot navigation, since their wider, sometimes global point of view can give instant information about the layout of a room, which onboard sensors would need multiple observations from different robot positions to recover, and they can be connected directly to power, avoiding drain on the robot's battery.

Human intuition suggests that navigating a robot from an exocentric viewpoint should be straightforward, as we can easily reason about spatial layouts from a third-person perspective and teleoperate robots from overhead views, but this intuition does not transfer to reinforcement learning within our setup. We find that PPO~\cite{schulman2017proximal}, which achieves near-perfect PointGoal navigation from onboard sensors~\cite{wijmans2019dd}, fails to learn navigation from exocentric images. The failure results from the perspective mismatch between the images, which are mostly agnostic to the robot's position and heading, and the egocentric actions and goal of DRL-based navigation, a mapping that proves intractable to resolve from the reward signal alone within our training budget and leaves the policy unable to reliably avoid obstacles.

Based on the findings about PPO, we present VINE, an agent composed of a supervised-learned neural network that estimates obstacle maps from exocentric camera images and a controller, either a navigation policy trained with PPO or a planner such as A*, that drives the robot towards a point goal using the estimated map. Our agent can navigate using cameras placed within wide ranges of positions without view-specific tuning, does not require intrinsics, and needs only an approximate camera pose, along with the robot pose and room dimensions. The obstacle map estimation network extends to setups where two cameras with partial views are used, and in all our comparisons it outperforms perspective mapping approaches. Despite operating from cameras randomized in position during evaluation, VINE surpasses the navigation success rate of PPO with an onboard camera.

\section{Related Work}

Anderson et al.~\cite{anderson2018evaluation} formalize embodied navigation into task categories, including point-goal navigation, where an agent must reach a target location specified as coordinates relative to its starting position. Wijmans et al.~\cite{wijmans2019dd} establish that end-to-end DRL, trained at scale over 2.5 billion frames from an onboard RGB-D camera, achieves near-perfect PointGoal navigation in unseen environments without explicit mapping or planning. Whether this success extends to exocentric cameras remains largely unexplored.

Few works have explored using exocentric cameras for robot navigation. Liu et al.~\cite{liu2021exploring} compare egocentric, exocentric, and combined perspectives for object navigation using Deep Q-Networks, but their setup uses a fixed exocentric camera in a single room, and limited training suggests the agent memorizes action sequences rather than learning a generalizable policy. Kim et al.~\cite{kim2024cctv} use fixed CCTV cameras to plan socially compliant paths in crowded environments, projecting the people detected in the camera images onto a 2D floor map through a homography transformation and feeding the result to a classical global planner. The approach assumes a known obstacle map of the static environment, and the homography must be computed for each camera, which demands map creation and camera calibration for every new scene. Our method does not assume a pre-built obstacle map, and it requires only approximate camera pose and no intrinsics.

Robinson et al.~\cite{robinson2026robot, robinson2026multi} control mobile robots from networks of uncalibrated CCTV cameras by planning in image space, but each camera requires its own trained robot detector and orientation estimator and a one-time segmentation of the drivable floor, which must be repeated whenever a camera is added or moved. VINE handles camera poses within its training range without per-camera training and estimates the obstacle layout from the images at every step.

Rather than learning navigation directly from exocentric images, VINE uses a learned obstacle map as an intermediate representation, following a line of work that decomposes visual navigation into spatial map construction and policy learning. Chaplot et al.~\cite{chaplot2020learning} decompose navigation into a learned module that constructs a top-down occupancy map and hierarchical policies for goal selection and action execution. This work demonstrates that intermediate spatial representations improve navigation over end-to-end approaches but assumes egocentric observations where the camera moves with the agent and does not address constructing spatial representations from static exocentric cameras at arbitrary viewpoints.

The problem of estimating a top-down occupancy map from camera images also arises in autonomous driving, where bird's-eye-view (BEV) representations are widely used. Philion and Fidler~\cite{philion2020lift} propose Lift-Splat-Shoot, which lifts image features into 3D frustums and splats them onto a BEV grid from an arbitrary number of cameras. The method is designed for camera rigs mounted on the vehicle. VINE instead works with exocentric cameras placed across wide ranges of positions and orientations and requires only approximate camera pose and no intrinsics.

In summary, prior work has decomposed visual navigation into map construction and planning, learned to estimate top-down maps from camera images, and used exocentric cameras to guide robots. To our knowledge, VINE is the first method to learn obstacle map estimation from exocentric cameras for robot navigation, producing an obstacle map that a learned policy or a classical planner can use directly. A single map estimator covers wide ranges of camera positions and orientations without per-camera training and requires only approximate camera pose and no intrinsics.

\section{Methodology}

We first define the navigation task and the simulation setup, then present VINE's architecture.

\subsection{Navigation Task}
\label{sec:task}

We consider PointGoal navigation~\cite{anderson2018evaluation}, in which a robot must reach a target position, specified as relative polar coordinates $(\cos\theta, \sin\theta, d)$ in the robot frame, without colliding with obstacles. The robot operates with a discrete action space $\mathcal{A} = \{\textit{forward},\, \textit{turn\_left},\, \textit{turn\_right}\}$, where the forward action moves the robot at $0.30\ \text{m/s}$ and the turning actions rotate it by $10^\circ$ per step at $5\ \text{Hz}$. An episode terminates when the robot reaches the goal (distance $\leq 0.2\ \text{m}$), collides with an obstacle or wall, or exceeds the time limit of $50\ \text{s}$.

At each step, the agent receives a state vector containing the goal position in polar coordinates $(\cos\theta, \sin\theta, d)$ and the robot's current forward and angular velocities, along with the robot pose $(x, y, \psi)$ and one or two camera images depending on the configuration. In exocentric configurations where camera pose varies across episodes, the agent additionally receives the camera pose (as a $4 \times 4$ matrix) and the room dimensions along the $x$ and $y$ axes.

\subsection{Simulation Setup}
\label{sec:sim}

We build our simulation setup on NVIDIA Isaac Lab~\cite{mittal2023orbit}, supporting obstacle and texture randomization and parallel environment instances. Rooms are rectangular with a fixed width of $6\ \text{m}$ and a variable depth of $3$--$6\ \text{m}$, and obstacle placement, room dimensions, and the robot and goal positions are randomized at every episode reset.

Our simulation setup has two rendering modes. The \textit{abstract} mode (Figure~\ref{fig:qualitative}a) uses procedurally generated rooms with randomly textured walls and floor and rectangular obstacles with cardboard box textures, enabling fast simulation for agent training and benchmarking. The photorealistic mode (Figure~\ref{fig:qualitative}b) uses detailed USD scenes with realistic furniture and materials, randomizing textures and positions while maintaining plausible furniture arrangements, and serves to validate that the approach transfers to visually complex environments once the obstacle map estimation network is trained on such environments. Both modes share the same parameters, differing only in the obstacles and their placement.

Table~\ref{tab:configs} summarizes the camera configurations used in our experiments. The No Camera configuration, which receives only the state vector, serves as a lower bound reflecting what can be achieved by following the goal reference while ignoring visual input entirely. The Fixed World-Mounted configuration uses a single overhead camera at a fixed pose and receives the robot pose in its state vector. Onboard images use $84 \times 84$ resolution following the Nature CNN convention~\cite{mnih2015human}, while exocentric cameras use higher resolution as they must resolve the full room layout, obstacles, and robot from a greater distance.

In the Random Camera Single configuration, a single camera is randomized per episode within $x \in [-2.0, 2.0] \text{m}$, $y \in [-0.8, 0.8] \text{m}$, $z \in [2.7, 2.9] \text{m}$, pointing toward the room center. We use domain randomization to simulate imperfect camera calibration in (a) the look-at target, perturbed by a uniform offset in $[-0.1, 0.1] \text{m}$ along both axes, (b) the intrinsics, varied through a zoom scale $s \sim \mathcal{U}(0.9, 1.0)$ yielding focal lengths in $[16.0, 16.9]\ \text{mm}$ and horizontal apertures in $[42.7, 45.0] \text{mm}$, (c) the camera pose matrix reported to the agent, corrupted with additive Gaussian noise ($\sigma = 0.075$, corresponding to a standard deviation of $7.5\ \text{cm}$ of translation error per axis and about $5^\circ$ of rotation error on average) applied independently to each of the 16 elements of the $4 \times 4$ matrix and sampled once per episode, and (d) the images, with Gaussian pixel noise whose standard deviation is sampled uniformly from $[0, 15]$ at every step.

The Random Camera Double configuration divides the room along the width axis, with each camera covering one half, $x_1 \in [0, 1.2] \text{m}$, $x_2 \in [-1.2, 0] \text{m}$, $y \in [-0.8, 0.8] \text{m}$, $z \in [2.2, 2.6] \text{m}$. Each camera's look-at target is shifted by $1\ \text{m}$ along the width axis toward its half, producing partial views with some overlap. The same domain-randomization scheme is applied (look-at perturbation, zoom, camera pose noise, and pixel noise), with focal lengths in $[9.0, 9.5] \text{mm}$ and horizontal apertures in $[22.8, 24.0] \text{mm}$ tuned so both cameras together maintain full environment coverage.

\begin{table}[htbp]
    \caption{Environment configurations.}
    \label{tab:configs}
    \centering
    \small
    \setlength{\tabcolsep}{3pt}
    \begin{tabular}{lccccc}
        \toprule
        Configuration & Cams & Res. & Cam.  & Robot & Placement \\
                      &      &      & pose  & pose  &           \\
        \midrule
        No Camera          & ---  & ---     & ---        & ---        & --- \\
        Onboard            & 1    & $84^2$  & ---        & ---        & On robot \\
        Fixed World-Mtd    & 1    & $256^2$ & ---        & \checkmark & Fixed (ext) \\
        Random Cam Single  & 1    & $256^2$ & \checkmark & \checkmark & Random (ext) \\
        Random Cam Double  & 2    & $256^2$ & \checkmark & \checkmark & Random (ext) \\
        \bottomrule
    \end{tabular}
\end{table}

\subsection{VINE}
\label{sec:vine}

At the exocentric image setting, standard PPO couples two different problems into a single objective, namely, extracting spatial structure from exocentric images and learning the mapping to egocentric actions, relying on reward signal alone to solve both. VINE decouples these two problems by lifting the perspective mapping out of the RL loop and reformulating it as a supervised pixel-to-pixel prediction problem, and therefore consists of two separate parts, an obstacle map estimation network trained with supervised learning and a controller that navigates from its output. The controller is one of two interchangeable options, either a policy trained with PPO, which acts on the obstacle map warped to the robot's egocentric frame, or a classical planner such as A*, which acts on the global obstacle map. We choose explicit obstacle map reconstruction over implicit latent alignment approaches (e.g., contrastive learning between exocentric and egocentric features) because the explicit map provides an interpretable intermediate representation whose quality can be measured directly via IoU, and because the controller does not need retraining when the perception module is improved or replaced, as we demonstrate in Section~\ref{sec:experiments}.

Formally, at each timestep the agent receives a state vector $\mathbf{s}_t$, the robot pose, and one or more camera images $\{I_t^i\}_{i=1}^{N}$, and selects an action $a_t \in \mathcal{A}$. VINE decomposes this into two stages with no shared gradients (Figure~\ref{fig:main}). First, the obstacle map estimation network $f_\theta$ maps the set of exocentric images and their corresponding camera poses to a global obstacle map $\hat{M}_t$. Second, the controller selects actions from this map, the robot pose, and the state vector. With the PPO controller, the policy $\pi_\phi$ receives the global map warped to the robot's frame, a viewpoint-invariant egocentric map $\mathbf{M}_t^{\text{ego}}$, and samples $a_t \sim \pi_\phi(\cdot \mid \mathbf{M}_t^{\text{ego}}, \mathbf{s}_t)$. With the A* controller, the planner instead plans a path to the goal on the global map $\hat{M}_t$ in the room frame, using the robot pose to locate the robot.

The obstacle map estimation network $f_\theta$ follows an encoder-decoder architecture with skip connections. It takes an exocentric image $I \in \mathbb{R}^{H \times W \times 3}$ and produces a global obstacle map $\hat{M} \in [0,1]^{64 \times 64}$, where each pixel encodes the probability of an obstacle at that spatial location over a fixed square region whose side equals the maximum room dimension. The map could have a higher resolution if needed, but this was enough for our environment size ranges. Since rooms vary in size, pixels outside the actual room bounds are excluded from the loss during training and set to $1.0$ (occupied) at inference. The module is conditioned on a vector $\mathbf{c}$ encoding the camera pose and room dimensions,
\[
    \mathbf{c} = \left[\, \tilde{\mathbf{t}},\; \text{vec}(\mathbf{R}),\; \tilde{\mathbf{a}} \,\right] \in \mathbb{R}^{14}
\]
where $\tilde{\mathbf{t}} \in \mathbb{R}^3$ is the normalized camera translation, $\text{vec}(\mathbf{R}) \in \mathbb{R}^9$ is the flattened rotation matrix, and $\tilde{\mathbf{a}} \in \mathbb{R}^2$ is the normalized room size along the $x$ and $y$ axes. This conditioning vector is injected via Feature-wise Linear Modulation (FiLM)~\cite{perez2018film}, which applies a learned affine transformation to intermediate feature maps,
\[
    \text{FiLM}(\mathbf{F}, \mathbf{c}) = \gamma(\mathbf{c}) \odot \mathbf{F} + \beta(\mathbf{c})
\]
where $\gamma$ and $\beta$ are learned functions of $\mathbf{c}$ and $\mathbf{F}$ denotes the feature map at a given layer, allowing the network to adapt its representations to each camera's viewpoint without requiring separate weights per camera. Camera intrinsics are not included in $\mathbf{c}$, the module handles intrinsic variation implicitly through domain randomization during training.

The encoder consists of the convolutional stages of a ResNet-34~\cite{he2016deep} pretrained on ImageNet, producing features at 64, 64, 128, 256, and 512 channels, with FiLM conditioning applied after each of the four residual stages, leaving the stem unmodulated. For multi-camera setups ($N > 1$), all cameras share encoder weights and are processed independently with their own conditioning vector $\mathbf{c}_i$. Per-camera features are then fused at every scale via attention-weighted summation, a two-layer MLP maps each camera's conditioning vector $\mathbf{c}_i$ to a scalar logit, and a softmax over cameras yields the fusion weights $w_i = \text{softmax}(\text{MLP}(\mathbf{c}_i))$. The weighted sum is refined by a $1 \times 1$ convolution with batch normalization, followed by Squeeze-and-Excitation channel attention~\cite{hu2018squeeze} (reduction ratio 16) to recalibrate feature importance. For single-camera setups the attention-weighted summation has nothing to combine and is skipped, while the refinement convolution and the channel attention still apply.

The decoder follows a UNet~\cite{ronneberger2015u} architecture with five upsampling blocks. Each block applies bilinear upsampling followed by a $3 \times 3$ convolution, concatenates the corresponding encoder skip connection, and processes the result through two further $3 \times 3$ convolutions with batch normalization, with spatial dropout ($p = 0.1$) after the first. The first four blocks take a skip connection, and the fifth, which returns the features to the input resolution, has none. FiLM conditioning is applied at each decoder block, after its last convolution, using the concatenated conditioning vectors of all cameras $[\mathbf{c}_1, \dots, \mathbf{c}_N]$. The final layer is a $1 \times 1$ convolution whose logits are resized bilinearly from the input resolution to $64 \times 64$ and passed through a sigmoid to give the obstacle map $\hat{M}$. The module is trained with a combined Dice and Binary Cross-Entropy loss,
\[
    \mathcal{L} = \lambda_{\text{bce}} \cdot \text{BCE}(\hat{M}, M) + (1 - \lambda_{\text{bce}}) \cdot \text{Dice}(\hat{M}, M)
\]
where $M$ is the ground-truth obstacle map and $\lambda_{\text{bce}} = 0.5$. Both terms are computed only over pixels inside the room bounds, and the BCE term is weighted by the ratio of free to occupied pixels to account for class imbalance. During training, color jitter (brightness, contrast, saturation, and hue) is applied to the input images.

For the PPO controller, the global obstacle map $\hat{M}$ is warped to the robot's egocentric frame via an affine transformation parameterized by the robot's pose $(x, y, \psi)$. We construct a $2 \times 3$ affine matrix applying rotation by $\psi$ and translation by $(x, y)$ in normalized map coordinates, and use bilinear grid sampling to produce $\mathbf{M}^{\text{ego}} \in [0,1]^{64 \times 64}$. Regions outside the global map are filled as occupied. This is a fixed geometric operation that solves the perspective mismatch before the obstacle map reaches the policy.

A policy $\pi_\phi$ is trained with PPO to select actions from the egocentric map $\mathbf{M}^{\text{ego}}$ and the state vector $\mathbf{s}_t$. The map is processed through a Nature CNN~\cite{mnih2015human} with three convolutional layers (32, 64, and 64 filters with kernel sizes 8, 4, and 3 and strides 4, 2, and 1), followed by a linear projection to a 256-dimensional feature vector. This vector is concatenated with $\mathbf{s}_t$ and fed to separate actor and critic heads, each a two-layer MLP with 128 units and Tanh activations. The actor outputs a categorical distribution over $\mathcal{A}$ and the critic estimates the state value.

The policy is trained to maximize the expected cumulative discounted reward $\mathbb{E}\left[\sum_{t=0}^{T} \gamma^t r_t\right]$, with a reward that combines four terms:
\[
    r_t = \alpha_p (d_{t-1} - d_t) + \alpha_s \mathbf{1}_{\text{success}} - \alpha_c \mathbf{1}_{\text{collision}} - \alpha_t
\]
where $d_t$ is the Euclidean distance between the robot and the goal at step $t$, and $\mathbf{1}_{\text{success}}$ and $\mathbf{1}_{\text{collision}}$ are indicator functions equal to $1$ at the step in which the robot reaches the goal or collides, and $0$ otherwise. The first term rewards progress toward the goal, the second provides a bonus upon reaching it, the third penalizes collisions, and the fourth is a small per-step cost that encourages efficient trajectories. We use $\alpha_p = 4.0$, $\alpha_s = 100.0$, $\alpha_c = 10.0$, and $\alpha_t = 0.01$, tuned by hand with the Onboard PPO baseline and copied over to all other configurations. The PPO baselines in Section~\ref{sec:experiments} use the same reward.

The A* controller plans on the global obstacle map $\hat{M}$ directly, using the robot pose to locate the robot on the map. Cells with an obstacle probability above $0.5$ are treated as obstacles, and each obstacle is grown by two cells ($0.19\ \text{m}$) so that the planned path keeps the robot's body clear of it. At every step, A* searches the 8-connected grid for the shortest path from the robot's cell to the goal's cell, and the robot follows it by heading toward the point on the path $0.2\ \text{m}$ ahead, turning when its heading differs from that direction by more than $15^\circ$ and moving forward otherwise. The planner uses the same three actions as the policy and needs no training.

\section{Experiments}
\label{sec:experiments}

We evaluate VINE in three stages. Section~\ref{sec:exp_occ} assesses the obstacle map estimation network in isolation, ablates its design choices and robustness to camera perturbations, and compares it against perspective mapping baselines. Section~\ref{sec:exp_nav} compares VINE, with both the PPO and A* controllers, against onboard and exocentric PPO baselines on the abstract environment. Section~\ref{sec:exp_photo} transfers VINE to photorealistic environments to verify that performance does not degrade severely on a more realistic setting.

\subsection{Obstacle Map Estimation}
\label{sec:exp_occ}

The obstacle map estimation network is trained on $80{,}000$ environment snapshots and evaluated on a held-out set of $10{,}000$. No room layout is shared between splits, verified via layout hashing. Each epoch subsamples $54{,}000$ training examples uniformly at random to reduce iteration time while preserving dataset diversity. Training runs for $150$ epochs with AdamW ($\text{lr} = 10^{-4}$, weight decay $10^{-4}$) and a ReduceLROnPlateau scheduler (factor $0.5$, patience $5$ epochs). The checkpoint with the highest validation IoU is retained, and the IoU we report is that of the final epoch.

Look-at perturbation is baked into the dataset at collection time, as it alters camera orientation and cannot be approximated post-hoc, while zoom scale, camera pose noise, pixel noise, and color jitter are re-sampled at every batch, with zoom approximated mathematically without re-rendering. The validation set is collected by varying camera parameters directly in Isaac Sim, providing a realistic test of generalization. IoU is computed only over pixels inside the room bounds.

Table~\ref{tab:occ_results} reports IoU for the Random Camera Single and Random Camera Double configurations. The network reaches $0.909$ validation IoU with one camera and $0.913$ with two, maintaining nearly identical accuracy despite the added complexity of fusing two partial views with independent viewpoints. Validation is performed under noisy camera poses, pixel noise, and varying intrinsics, yet the model receives no intrinsic parameters as input, which demonstrates robustness to both camera pose perturbations and intrinsic variation. Figure~\ref{fig:qualitative} shows qualitative predictions across diverse camera placements and room layouts.

\begin{table}[htbp]
    \caption{Obstacle map estimation IoU in train and validation sets.}
    \label{tab:occ_results}
    \centering
    \small
    \setlength{\tabcolsep}{4pt}
    \begin{tabular}{lcc}
        \toprule
        Configuration & Train IoU & Val IoU \\
        \midrule
        Random Cam Single & 0.929$\pm$0.035 & 0.909$\pm$0.015 \\
        Random Cam Double & 0.947$\pm$0.008 & 0.913$\pm$0.007 \\
        \bottomrule
    \end{tabular}
\end{table}

\textbf{Perspective mapping:} We compare the network against three baselines that do not learn the mapping, and instead project the image onto the floor using the camera pose and a pinhole camera model. This projection needs the camera intrinsics, which VINE does not receive. We give the baselines the exact intrinsics on the unperturbed sets, and the intrinsics at the middle of the zoom range on the validation sets, since the zoom of each image is not known there. The camera pose noise of the validation sets makes the rotation part of the pose matrix invalid, so the baselines replace it with the closest valid rotation. All three baselines first segment the obstacles in the image. We detect them with Grounding DINO~\cite{liu2024grounding} from text prompts and pass each detection to SAM~2.1~\cite{ravi2025sam}, which returns one mask per obstacle. The prompts and detection thresholds were tuned on training images, where they find $98\%$ of the obstacles.

The baselines differ in how they take these masks to the floor. Contact hull assumes that the lowest pixel of a mask in each image column is a point where the obstacle touches the floor. We cast a ray from the camera through each of these pixels, find where it hits the floor, and fill the convex hull of the resulting points. Since the camera only sees the bottom edges that face it, the hull covers only part of the obstacle. Carving works in the opposite direction, and projects the center of each map cell into the image. A cell is marked as occupied when it falls inside a mask in every camera that sees it. The sides and top of an obstacle also cover the floor behind it in the image, so carving marks more cells than the obstacle occupies. Depth uses Depth Anything V2~\cite{yang2024depth} to recover the 3D position of the mask pixels. The model predicts depth up to an unknown scale and shift, which we fit on the floor pixels, since their real depth is known from the camera pose. We then keep the points higher than $8\ \text{cm}$ above the floor, fill their convex hull for each obstacle, and intersect the result with carving.

Table~\ref{tab:persp} reports IoU on the validation sets and on $2{,}000$-sample sets without camera perturbations. Without perturbations, the best baseline reaches $0.80$ against $0.91$ for VINE, as the cameras see only the visible faces of each obstacle, and under camera pose noise every projection shifts with the pose error, dropping the baselines to $0.53$ or below while VINE keeps its accuracy.

\begin{table}[htbp]
    \caption{Comparison with perspective mapping (IoU).}
    \label{tab:persp}
    \centering
    \small
    \setlength{\tabcolsep}{4pt}
    \begin{tabular}{lcccc}
        \toprule
        Method & \multicolumn{2}{c}{Unperturbed} & \multicolumn{2}{c}{Validation} \\
        \cmidrule(lr){2-3} \cmidrule(lr){4-5}
               & Single & Double & Single & Double \\
        \midrule
        Contact hull & 0.554 & 0.605 & 0.405 & 0.450 \\
        Carving & 0.708 & 0.731 & 0.511 & 0.500 \\
        Depth & 0.796 & 0.800 & 0.525 & 0.511 \\
        \midrule
        VINE & 0.914 & 0.919 & 0.909 & 0.913 \\
        \bottomrule
    \end{tabular}
\end{table}

Next, we ablate the two major design choices in the supervised obstacle map estimation network, namely, FiLM conditioning and domain randomization. We then isolate individual perturbation types to analyse their effect on prediction accuracy. All ablations use the Random Camera Single configuration.

\textbf{FiLM conditioning:} We train a variant in which all FiLM layers are replaced by identity operations ($\gamma = 1$, $\beta = 0$) for $100$ epochs, removing the influence of the camera pose vector on intermediate features. All other hyperparameters and data remain identical. Validation IoU drops from $0.909$ to $0.592$, confirming that the module needs explicit camera pose conditioning to generalize reliably across viewpoints. Without access to camera pose, the network must infer the room's orientation from the image alone, which is unlikely in completely randomized rooms such as the ones in our environment. The predictions collapse into blobs without commitment to any specific layout.

\textbf{Domain randomization:} We train a second variant with all on-the-fly augmentations disabled (zoom, camera pose noise, pixel noise, and color jitter), while look-at perturbation remains as it is baked into the dataset at collection time. Validation IoU peaks within the first four epochs, at $0.626$ on average, and then declines while training IoU keeps rising to $0.898$, ending at $0.570$ compared to $0.909$ for the full model. The gap confirms that training-time domain randomization is essential to avoid overfitting and to make the model robust to the perturbations encountered at deployment.

\begin{figure*}[htbp]
\centering
\includegraphics[width=0.8\linewidth]{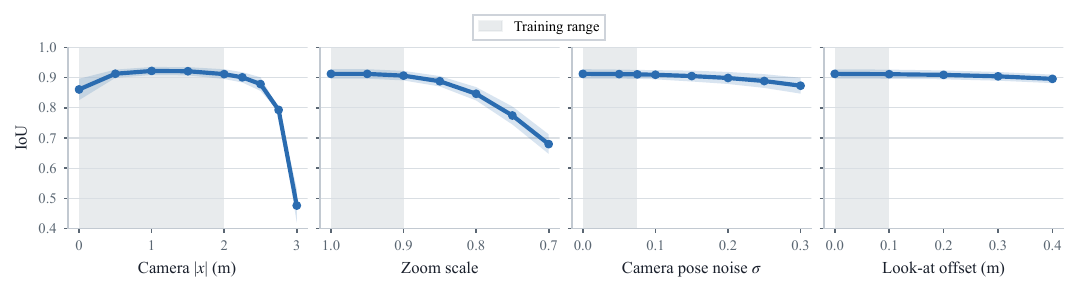}
\caption{Obstacle map estimation IoU as a function of individual perturbation magnitude. Lines show the mean over three estimator seeds and bands the $95\%$ confidence interval. Gray regions indicate the training range.}
\label{fig:sensitivity}
\end{figure*}

\textbf{Sensitivity analysis:} We evaluate the trained single-camera model on 1,000-observation validation sets with unseen obstacle layouts under controlled perturbation levels. For each perturbation type, we sweep from within the training range to beyond it while zeroing all other noise sources. Camera position varies within its training range across all sweeps. Figure~\ref{fig:sensitivity} reports the results.

IoU degrades gracefully beyond the training distribution for most perturbations. Camera pose noise lowers IoU from $0.912$ to $0.873$ at $4\times$ the training level, and look-at offset keeps it above $0.90$ up to $3\times$ the training range, which indicates tolerance to camera pose errors well beyond those seen during training. Zoom holds $0.89$ at $0.85$ before falling to $0.68$ at $0.7$. Camera position is the only exception, as IoU holds at $0.88$ at $|x| = 2.5\text{ m}$, beyond the $2.0\text{ m}$ training range, and then drops sharply to $0.48$ at $3.0\text{ m}$.

\subsection{Navigation Performance}
\label{sec:exp_nav}

All RL experiments use PPO with the Adam optimizer~\cite{kingma2014adam}. The learning rate decays linearly from $3 \times 10^{-4}$ to $5 \times 10^{-5}$, the clipping parameter from $0.2$ to $0.1$, and the entropy coefficient from $0.05$ to $0.005$. We set the discount factor $\gamma = 0.995$, GAE~\cite{schulman2015high} parameter $\lambda = 0.95$, value function coefficient $0.5$, gradient norm clipping at $0.5$, and early stopping when the approximate KL divergence exceeds $0.045$. Each update uses $4$ epochs over a rollout buffer of $64$ steps $\times$ $128$ parallel environments ($8{,}192$ transitions) with minibatches of $4{,}096$. These hyperparameters were tuned on the Onboard PPO baseline and are shared across all configurations, including the PPO controller of VINE.

Each configuration is trained for $20$M steps across $128$ parallel environment instances. Evaluation runs a deterministic policy for $10{,}000$ episodes distributed evenly across $128$ environments, preventing short episodes from biasing the distribution. We report success rate (robot reaches the goal without colliding with obstacles), collision rate (robot collides with an obstacle, failing the task), and time-limit rate (robot does not collide but does not reach the goal). Every configuration is run for three seeds, and we report the mean with a $95\%$ confidence interval over those seeds.

Table~\ref{tab:nav_results} presents the results. The No Camera baseline receives only the state vector and succeeds in $25.8\%$ of episodes, establishing the rate at which the goal can be reached by following the reference direction with no obstacle awareness. PPO with a Fixed World-Mounted exocentric camera achieves $24.7\%$, comparable to the No Camera configuration. The policy does not learn to solve the perspective transformation between exocentric images and egocentric actions under the training budget, and thus collides or exceeds time limit most of the time.

VINE with the PPO controller and Random Camera Single achieves $92.3\%$ success, above Onboard PPO at $83.9\%$, with non-overlapping confidence intervals. This indicates that, in the proposed setup, the predicted obstacle map provides spatial information  equivalent or better than that of a robot-mounted camera for the purpose of obstacle avoidance and goal-directed navigation. With Random Camera Double, VINE achieves $93.5\%$, showing that fusing two partial views introduces no measurable degradation relative to the single-camera setup.

With the A* controller, VINE achieves $94.4\%$ success with one camera and $95.1\%$ with two, with confidence intervals overlapping those of the PPO controller, which shows that VINE does not depend on a specific controller. A* collides more often than PPO but almost never reaches the time limit, and it succeeds in $99.9\%$ of episodes on the ground truth map, so nearly all of its failures come from errors in the estimated map and trajectories that pass too close to the obstacles.

\begin{table}[htbp]
    \caption{Navigation performance across configurations.}
    \label{tab:nav_results}
    \centering
    \small
    \setlength{\tabcolsep}{4pt}
    \begin{tabular}{lccc}
        \toprule
        Configuration & Succ. & Coll. & T.L. \\
                      & (\%)  & (\%)  & (\%) \\
        \midrule
        No Camera PPO & 25.8$\pm$2.6 & 74.2$\pm$2.6 & 0.0$\pm$0.1 \\
        Onboard PPO & 83.9$\pm$4.7 & 6.0$\pm$6.3 & 10.2$\pm$5.5 \\
        Fixed World-Mtd PPO & 24.7$\pm$2.3 & 64.8$\pm$38.5 & 10.5$\pm$36.3 \\
        \midrule
        VINE PPO, Single & 92.3$\pm$3.1 & 3.4$\pm$1.7 & 4.3$\pm$3.6 \\
        VINE PPO, Double & 93.5$\pm$1.5 & 3.0$\pm$1.9 & 3.5$\pm$1.2 \\
        VINE A*, Single & 94.4$\pm$2.2 & 5.0$\pm$2.0 & 0.7$\pm$0.3 \\
        VINE A*, Double & 95.1$\pm$1.2 & 4.4$\pm$0.9 & 0.5$\pm$0.3 \\
        \midrule
        A*, GT map & 99.9$\pm$0.1 & 0.1$\pm$0.1 & 0.0$\pm$0.1 \\
        \bottomrule
    \end{tabular}
\end{table}

\begin{figure*}[htbp]
\centering
\includegraphics[width=\linewidth]{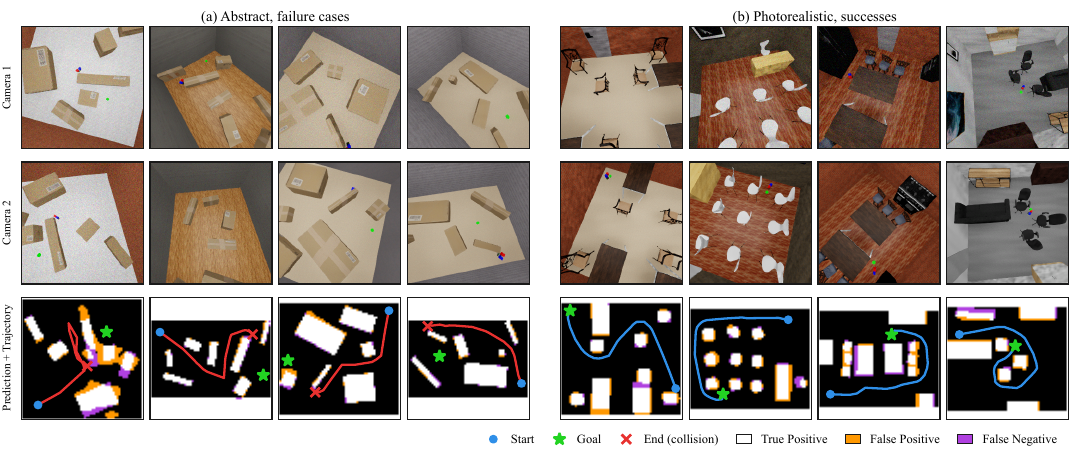}
\caption{Qualitative results of VINE with the PPO controller in the Random Camera Double setup. Each column shows the two camera views and the predicted obstacle map with the driven trajectory, where colored pixels indicate prediction errors. (a) Failure cases in the abstract environment. (b) Successful episodes in photorealistic environments, where the last column is a held-out room with furniture unseen during training.}
\label{fig:qualitative}
\end{figure*}

Figure~\ref{fig:qualitative}a shows failure episodes for VINE with the PPO controller in the Random Camera Double configuration. Prediction errors are small in most of the shown cases. Failures occur in environments where the robot passes through narrow passages between obstacles, leading to collisions, and are therefore attributable to map estimation errors and the RL policy taking the robot too close to obstacles.

\subsection{Photorealistic Environments}
\label{sec:exp_photo}

To verify that VINE does not depend on the visual simplicity of abstract rendering, we evaluate it in photorealistic environments with realistic furniture, materials, and lighting. We define $204$ room environments partitioned into three splits. The \textbf{supervised} split ($180$ environments) is used for obstacle map estimation network training and uses completely random furniture arrangements to reduce overfitting. The \textbf{known obstacles} split ($20$ environments) evaluates VINE on new plausible layouts, randomized per episode, with the furniture seen during supervised training. The \textbf{held-out} split ($4$ environments) contains $13$ furniture items that never appear in the supervised split, testing VINE under unknown obstacle types. All rooms have a fixed width of $6\text{ m}$ and depth between $3\text{ m}$ and $6\text{ m}$. This experiment uses the Random Camera Double configuration with the same domain randomization as the abstract setup.

Table~\ref{tab:photo_occ} reports obstacle map estimation IoU across splits. The network generalizes well to unseen room layouts, but the $103$ furniture items seen during training are not sufficient to retain full accuracy on the held-out items. As the held-out column of Figure~\ref{fig:qualitative}b shows, the obstacle positions are still accurately localized in the held-out split, although with higher errors than known obstacles.

\begin{table}[htbp]
    \caption{Obstacle map estimation IoU in photorealistic environments.}
    \label{tab:photo_occ}
    \centering
    \small
    \setlength{\tabcolsep}{4pt}
    \begin{tabular}{lc}
        \toprule
        Split & IoU \\
        \midrule
        Supervised (180 envs) & 0.908$\pm$0.016 \\
        Known obstacles (20 envs) & 0.840$\pm$0.004 \\
        Held-out obstacles (4 envs) & 0.807$\pm$0.008 \\
        \bottomrule
    \end{tabular}
\end{table}

Rather than retraining PPO, we directly transfer the three policies trained in the abstract Random Camera Double configuration and evaluate them on the known obstacles and held-out splits. Each policy is evaluated for $2{,}000$ episodes on the known split and $500$ on the held-out split, totaling $6{,}000$ and $1{,}500$ episodes across the three seeds. Table~\ref{tab:photo_nav} reports navigation performance over those seeds. VINE achieves $90.8\%$ success on the known obstacle environments and $77.2\%$ on the held-out set, without any retraining or fine-tuning of the policy, as once the obstacle map estimation network adapts to a new visual domain, the policy transfers without modification. The performance gap relative to the abstract environment can be attributed to room configurations containing obstacles with different sizes than those encountered during training, causing the policy to collide more often.

\begin{table}[htbp]
    \caption{Navigation in photorealistic environments (Random Cam Double).}
    \label{tab:photo_nav}
    \centering
    \small
    \setlength{\tabcolsep}{4pt}
    \begin{tabular}{lccc}
        \toprule
        Split & Succ. & Coll. & T.L. \\
              & (\%)  & (\%)  & (\%) \\
        \midrule
        Known (20 envs) & 90.8$\pm$2.0 & 5.2$\pm$2.7 & 4.0$\pm$0.7 \\
        Held-out (4 envs) & 77.2$\pm$2.0 & 17.7$\pm$0.8 & 5.1$\pm$1.3 \\
        \bottomrule
    \end{tabular}
\end{table}

Our experiments demonstrated that the method is not only capable of navigating robots in indoor environments from external cameras but also that its map-estimation module is highly robust. The decoupled architecture offers practical advantages. Because perception and control do not share gradients during training, the RL policy requires no retraining when the obstacle map estimation network is improved or adapted to a new visual domain, as demonstrated by the direct transfer to photorealistic environments. This separation also simplifies sim-to-real transfer, reducing it from a reinforcement learning problem to a computer vision problem where established techniques apply directly. 

The main limitation is the requirement for known robot pose, camera pose, and room dimensions at inference time, along with the rectangular room assumption. Predicting the egocentric obstacle map directly in the robot frame could bypass the need for such information. Alternatively, a pose estimation network trained in the same supervised fashion as the obstacle map estimation network could predict the robot's pose from exocentric images and camera pose, removing the dependence on external localization. Another limitation is the case when an obstacle cannot be seen in the images due to other obstacles occluding it. We leave improvements on the discussed limitations and real robot experiments for future work.

\section{Conclusion}

VINE shows that exocentric cameras can replace onboard sensors for robot navigation when the perspective mismatch between camera-frame observations and egocentric actions is resolved at the perception level rather than learned from reward. The agent generalizes across randomized camera poses, extends to multi-camera setups, and is robust to approximate calibration and intrinsic variation. Infrastructure-mounted cameras represent a practical and largely under-explored sensing modality for indoor robot navigation.

\section{Acknowledgments}

The Large Language Model Claude Opus from Anthropic (on multiple versions) was used for coding, code review, and improving writing (initially sketched by the authors). The authors have reviewed its work thoroughly.

\bibliographystyle{IEEEtran}
\bibliography{ref}

\end{document}